# Classification of Potholes Based on Surface Area Using Pre-Trained Models of Convolutional Neural Network

Chauhdary Fazeel Ahmad [1], Abdullah Cheema [2], Waqas Qayyum [3], Rana Ehtisham [4], Muhammad Haroon Yousaf [5], Junaid Mir[6] , Nasim Shakouri Mahmoudabadi[7]and Afaq Ahmad [8]

[1]    University of Engineering and Technology Taxila, Rawalpindi, Pakistan; 18-CE-61@students.uettaxila.edu.pk;
[2]    University of Engineering and Technology Taxila, Rawalpindi, Pakistan;. 18-CE-49@students.uettaxila.edu.pk;
[3]    University of Engineering and Technology Taxila, Rawalpindi, Pakistan; waqas.qayyum@students.uettaxila.edu.pk.
[4]    University of Engineering and Technology Taxila, Rawalpindi, Pakistan; rana.ehtisham@students.uettaxila.edu.pk.
[5]    University of Engineering and Technology Taxila, Rawalpindi, Pakistan; haroon.yousaf@uettaxila.edu.pk
[6]    University of Engineering and Technology Taxila, Rawalpindi, Pakistan; junaid.mir@uettaxila.edu.pk.
[7]    Department of Civil Engineering, The University of Memphis, TN 38152, USA; nshkrmhm@memphis.edu;
[8]    University of Memphis, Memphis TN USA aahmad4@memphis.edu
*    Correspondence: aahmad4@memphis.edu

**Abstract:** Potholes are fatal and can cause severe damage to vehicles as well as can cause deadly accidents. In South Asian countries, pavement distresses are the primary cause due to poor subgrade conditions, lack of subsurface drainage, and excessive rainfalls. The present research compares the performance of three pre-trained Convolutional Neural Network (CNN) models, i.e., ResNet 50, ResNet 18, and MobileNet. At first, pavement images are classified to find whether images contain potholes, i.e., Potholes or Normal. Secondly, pavements images are classified into three categories, i.e., Small Pothole, Large Pothole, and Normal. Pavement images are taken from 3.5 feet (waist height) and 2 feet. MobileNet v2 has an accuracy of 98% for detecting a pothole. The classification of images taken at the height of 2 feet has an accuracy value of 87.33%, 88.67%, and 92% for classifying the large, small, and normal pavement, respectively. Similarly, the classification of the images taken from full of waist (FFW) height has an accuracy value of 98.67%, 98.67%, and 100%.

## 1. Introduction

Roads are crucial for transportation in any country. They connect people, move goods, and help businesses grow. In places like Pakistan and other South Asian countries, heavy rainy seasons can lead to widespread damage to road surfaces. Roads contribute significantly to an economy's overall growth. Asphalt and concrete are commonly used materials in road construction. Potholes are bowl-shaped holes in the road surface that have a distinct texture. Low temperatures amplify the formation of potholes because water expands when it freezes to form ice, placing additional strain on a cracked pavement or road. Once a pothole has formed, it expands by removing pavement fragments. If a pothole fills with water, its growth may be accelerated because the water "washes away" loose road surface particles as vehicles pass over it. Potholes can grow to a width of several feet, but their depth rarely exceeds a few inches. If they grow large enough, tires and vehicle suspensions are damaged. This can directly lead to severe road accidents, especially on highways where vehicle speeds are higher. One of the processes of pothole formation is shown in Figure 1. Potholes can also (less frequently) result from a variety of non-structural causes, such as diesel (or other chemicals) spillages, mechanical damage to road surfaces from vehicle rims, accidents, and fires, damage caused by falling rocks in cuttings, animal hooves on the road surface in hot weather, Poor road design over specific subgrades (expanding, collapsing, dispersing), lack of bond between the surfacing and WBM base, and insufficient bitumen [1].

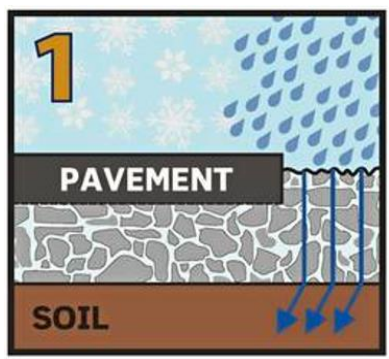

1..Potholes begin after snow or rain seeps into the soil below the road surface.

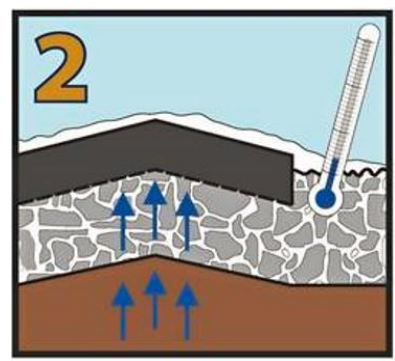

2. The moisture freezes when temperatures drop, causing the ground to expand and push the pavement up

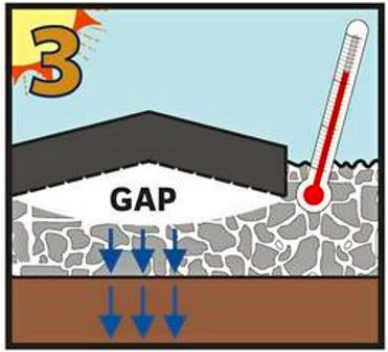 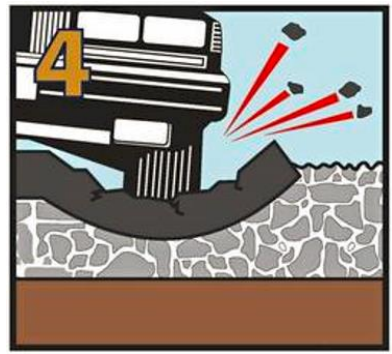

3. As the temperature rises, the ground returns to normal, but the pavement often remains raised. This creates a gap between the pavement and the ground below it.

4. When vehicles drive over this cavity, the pavement surface cracks fall into the hollow space leading to the birth of another pothole.

**Figure 1:** Stages of Pothole Formation: From Moisture Infiltration to pothole birth [1]

Pothole detection is essential to road maintenance and repair because of its negative impact on road quality and vehicle safety [2]. Potholes developed due to poor materials, a poor design that causes surface water to accumulate, and ice formation in cracks. Potholes affect the life of people and property every year [3]. Since 2011, for five years, drivers have spent more than $3 billion on cars to fix damage caused by potholes. The approximate cost per driver comes up to $300 [4]. According to an India Economic Times story from 2018 [5], the Supreme Court found 3597 casualties related to highway potholes. Potholes were India's leading causes of accidents in 2019, accounting for nearly 4,770 incidents. World Health Organization (WHO) states that in 2010 1.25 million people died from road accidents [6].

In Deep learning (DL) methods, CNN is the most commonly used algorithm. One of the important benefits of CNN is that it automatically detects the relevant features without human supervision [7]. CNN is widely used in Computer Vision [8], speech processing [9], and Face recognition [10]. Commonly, CNN is composed of multiple layers, including convolutional layers, pooling layers, and fully connected (FC) layers [11]. An example of CNN architecture for detecting the pothole is shown in Figure 2. CNN layers classify pothole and normal pavement images by progressively extracting hierarchical features through convolutional and pooling layers, enabling the network to recognize intricate patterns and structures that contribute to accurate image classification. One important approach in Machine Learning (ML) is transfer learning, in which models developed for one task are used for another task. There are several pre-trained CNN models which includes Alexnet [12], Googlenet [13], Inception v3 [14], Densenet 201 [15], Darknet (19 and 53) [16], Resnet (18,50 and 101) [17], Vgg (16 and 19) [18] [19], Shufflenet [20], Nasnet (mobile and large) [21], Mobilenet v2 [22], Xception [23], Inceptionresnet v2 [24], Squeezenet [25] etc.

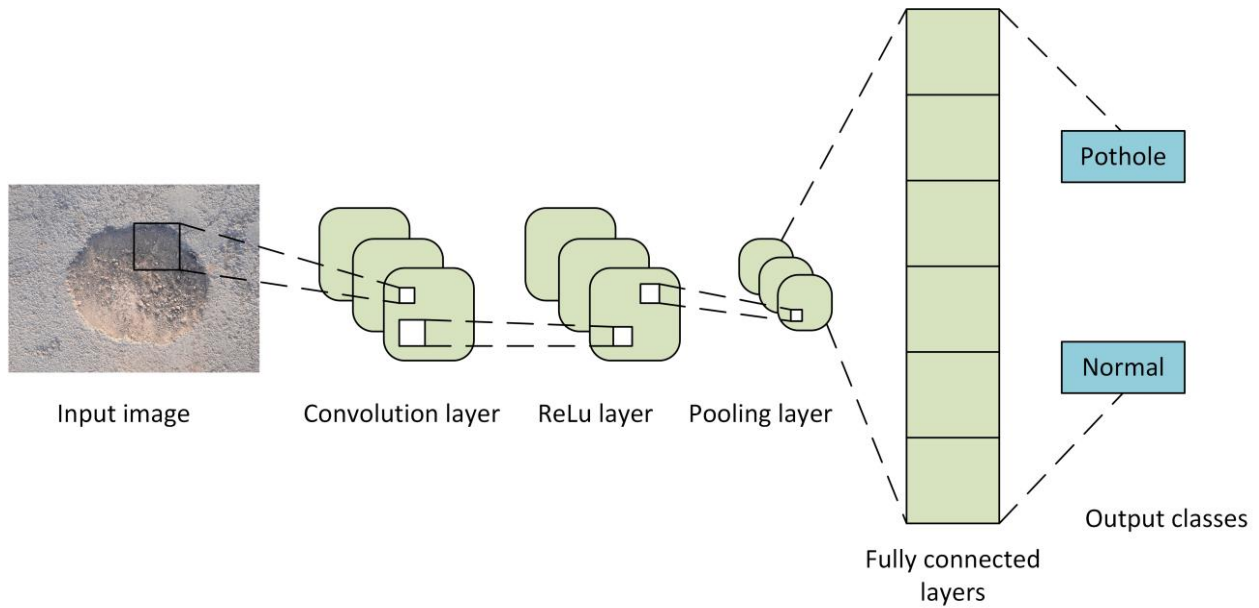

**Figure 2:** Example of classification with CNN

Figure 3 shows the CNN architecture categorizing potholes into the small, large, and normal pavement. CNN is widely utilized for detecting various types of civil engineering structure damage. The damages include cracks, corrosion, and potholes, among others. Previously, many researchers have used computer vision and ML to detect various types of structural damage. Section 2 (review of the literature) summarized the work of various researchers.

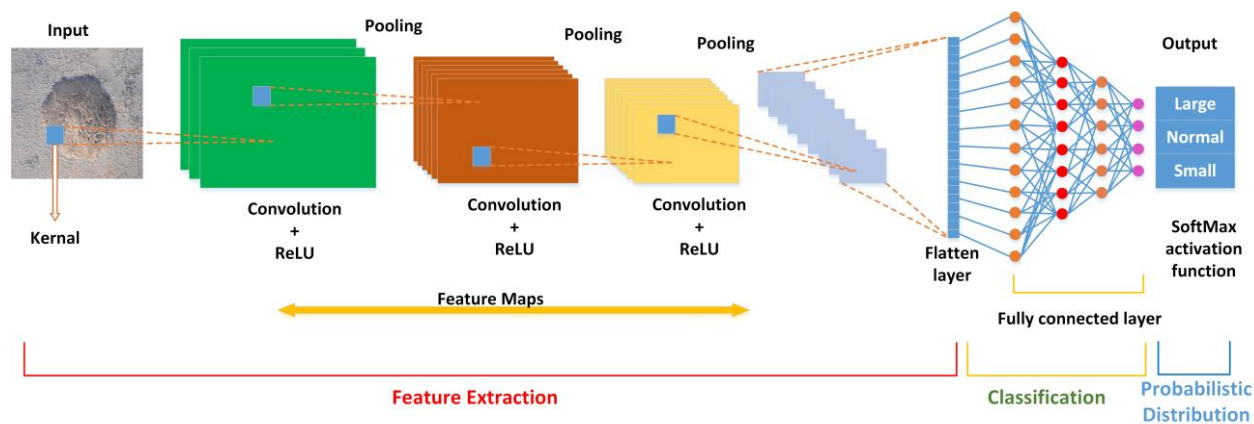

**Figure 3:** Architecture of CNN

Accurately detecting potholes is an essential step in determining the best methods for asphalt-surfaced pavement maintenance and rehabilitation because they are indicators of underlying structural defects in the road [26]. Potholes must be detected before they take any person's life or cause more significant damage. The potholes are of two main types one which doesn't have any effect on the vehicle and one which has an effect on the vehicle or whose surface area is greater than the contact area of the tire of the car. Table 1 shows the detail of the tire's contact area.

**Table 1:** Tires contact pressure based on imprint shapes [27]

| Imprint shape | Wheel pressure (MPa) | Contact area (sq. mm) |
|---|---|---|
| Circular | 0.68 | 60000 (93 sq. in) |
| Rectangular | 0.68 | 61575 (95 sq. in) |
| Ellipse | 0.68 | 60416 (93.6 sq. in) |
| Actual | 0.68 | 60318 (93.5 sq. in) |

In this study, three types of classification were performed. The first classification includes two categories, Normal and Pothole, without considering the height from which the images were taken. In the Second classification, images taken from the height of 2 feet were classified into three categories, Normal, Small, and Large. Normal corresponds to the fact that there is no pothole in the pavement; small is for the pavement image of the pothole with an area less than the contact area of the tires, and Large is for the pavement with a pothole area greater than the contact area of the tires (93 sq. in, shown in table 1). In the third classification, images taken from the FFW were classified into three categories, Normal, Small, and Large. The study sections are as follows: Literature Review is given in Sect. 2, Proposed Methodology in Section 3, Experimentation and Results in Section 4, and Conclusion in Section 5.

## 2. Literature Review

Previously, different researchers had done much research on the automatic detection of potholes. Abhinav and Priti [28] proposed a method for pothole detection based on CNN; the model's validation accuracy was 92%, while the testing accuracy was 80 %. In [29], the authors classified the road pavements based on area, depth, and volume. They used a specific mobile application called Sketches and Calc to calculate area and volume, and on that base, they classified the road pavements into different categories. In [30], a logical strategy for identifying and locating a pothole in the road was proposed. The data was collected using smartphone sensors, making the method quick and economical. Class imbalance categorization is simplified using a basic yet functional Neural Network, improving the accuracy of previously reported measurements. Accuracy of 94% and considerable recall of 81% make this model acceptable for real-time applications. Taehyeong et al. [31] examine the work of many researchers on pothole identification utilizing vibration-based approaches, 3D reconstruction-based methods, and vision-based methods. Lightweight one-stage object recognition network, as suggested by Zhang et al. [32] To guarantee precision and timeliness in sinkhole identification, they built a network with a custom LF (lightweight feature extraction) module and a NAM (Normalization-based Attention Module) attention module. By fusing CSPDarknet53-tiny with spatial pyramid pooling and feature pyramid network, Heo et al. [33] created SPFPN-YOLOv4 tiny. The dearth of data was made up for by using data enrichment techniques such as gamma regulation, horizontal flipping, and scaling to acquire a total of 2665 datasets, which were then split into training, validation, and test groups at 70%, 20%, and 10%, respectively. When the mean average accuracy (intersection over union = 0.5) of YOLOv2, YOLOv3, YOLOv4, and SPFPN-YOLOv4 tiny was compared, the SPFPN-YOLOv4 tiny was found to perform roughly 2-5% better. Table 2 summarizes the work of different researchers for the detection of potholes.

**Table 2:** Different researchers work on detection of pothole

| References | Methods/ Algorithms | Input | Results |
|---|---|---|---|
| | | | |
| Koch and Brilakis [34] | image pro-cessing | RGB image | Accuracy = 85.9% <br> Precision = 81.6% <br> Recall = 86.1% |
| Ryu et al. [35] | image pro-cessing | RGB image | Accuracy = 91.0% <br> Precision = 85.3% <br> Recall = 93.5% |
| Schiopu et al. [36] | image pro-cessing | RGB image | Precision 90% <br> Recall 100% |
| Jak_stys et al. [37] | image pro-cessing | RGB image | Accuracy = 78% |
| Akagic et al. [38] | image pro-cessing | RGB image | Accuracy = 82% |
| Wang et al. [39] | image pro-cessing | Grayscale image | Pothole detection: <br> Accuracy = 86.7% <br> Precision = 83.3% <br> Recall = 87.5% <br> Pothole segmentation: <br> Accuracy= 88.6% |
| Fan et al. [40] | image pro-cessing | Transformed disparity image | Road damage detection <br> Accuracy = 97.56% |
| Fan et al. [41] | image pro-cessing | Transformed disparity image | Accuracy = 98.7% <br> F1-score = 89.4%. |
| Zhang [42] | 3-D point cloud / Stereo vision, quadratic surface fitting, connected component la-belling (CCL) | | Absolute disparity error = 6.82% |
| Li et al. [43] | stereo vision system | | Pothole detection time = 5s, <br> Pothole detection range = 14m in advance. |
| Daniel and Preeja [44] | Image classifi-cation / SVM | Grayscale image | The error rate for detecting cracks and potholes <br> Test 1 = 15% <br> Test 2 = 13.5% <br> Test 3 = 14% |
| Hadjidemetriou et al. [45] | Image classifi-cation / SVM, DCT, GLCM | Grayscale image | Accuracy = 81.97 % <br> Precision = 64.21 % <br> Recall = 91.21 %. |
| Hoang [46] | Image classifi-cation/LS-SVM, ANN | Grayscale image | LS-SVM and ANN <br> Accuracy = 85% <br> LS-SVM <br> Accuracy = 89% |
| Pan et al. [47] | Image classifi-cation / ANN, RF, SVM | RGB image, Multi-spectral Image | With the UAV MSI Accuracy = 98.3% |
| Gao et al. [48] | Image classifi-cation / LIBSVM | RGB image, | Recall = 100% Precision = 97.4% F1-Score = 98.7% |
| Pereira et al. [49] | Image classifi-cation / Self-de-signed DCNN | RGB image, | Accuracy = 99.80 % Precision = 100% Recall = 99.60% F1-Score = 99.60% |
| An et al. [50] | Image classifi-cation / Incep-tion_ResNet_v2, ResNet_v2 -151, and Mo-bileNet_v1 | RGB image, Grayscale image | Accuracy = 96.5~97.5% |
| Ye et al. [51] | Image classifi-cation / Self-de-signed DCNN | RGB image | Precision = 98.95% |
| Bhatia et al. [52] | Image classifi-cation / Self-de-signed DCNN | Thermal image | Accuracy = 97.08% |
| Suong et al. [53] | Object detection / YOLO | RGB image | Precision = 60.14% to 82.43% |
| Maeda et al. [54] | Object detection / SSD | RGB image | Recalls and precision greater than 75% with an in-ference time of 1.5sec on the smartphone. |
| Wang et al. [55] | Object detection / Faster R-CNN | RGB image | Mean F1-Score = 62.55 % |
| Ukhwah et al. [56] | Object detection / YOLO | Grayscale image | Mean Average Precision Yolo v3 = 83.43% Yolo v3 Tiny = 79.33% Yolo v3 SPP = 88.93% Area Measurement Accu-racy Yolo v3 = 64.45% Yolo v3 Tiny = 53.26% Yolo v3 SPP = 72.10% |
| Dharneeshkar et al. [57] | Object detection / YOLO | RGB image | Yolo v3 (416 x 416) Precision = 69% Recall = 35% Yolo v2 (416 x 416) |

| | | | Precision = 46% |
|---|---|---|---|
| | | | Recall = 45% |
| | | | Yolo v2 (608 x 608) |
| | | | Precision = 52% |
| | | | Recall = 50% |
| | | | Tiny Yolo v3 (608 x 608) |
| | | | Precision = 76% |
| | | | Recall = 40% |
| Baek and Chung [58] | Object detection / YOLO | RGB image | Precision = 83.45<br>Accuracy = 77.86 |
| Kortmann et al. [59] | Object detection / Faster R-CNN | RGB image | F1score = 48.7% |
| Yebes et al. [60] | Object detection / Faster R-CNN, SSD | RGB image | Mean average precision above 75%<br>5-6 frames per second<br>Speeds below 60 km/h |
| Gupta et al. [61] | Object detection / SSD | Thermal image | modified ResNet34<br>Precision = 74.53%<br>modified ResNet50-Reti-naNet<br>Precision = 91.15% |
| Javed et al. [62] | Object detection / R-CNN, SSD | RGB image | Accuracy = 93% |
| Pereira et al.[63] | Semantic segmentation / U-Net | RGB image | Accuracy greater than 97% mean Intersection Over Union (mIoU) = 0.86 |
| Chun and Ryu [64] | Semantic segmentation / FCN | RGB image | Supervised<br>Precision = 70.12%<br>Recall = 92.64%<br>Accuracy = 87.28%<br>F1-Score = 79.82%<br>Semi-supervised<br>Precision = 90.14%<br>Recall = 88.22%<br>Accuracy = 93.87%<br>F1-Score = 89.17% |
| Fan et al. [65] | Semantic segmentation / AA, GAN | RGB image, transformed disparity image | AA-UNet mIoU, 14% compared with the SoA single-modal networks,<br>AA-RTFNet mIoU, 8% compared with the SoA data-fusion networks. |

| Masihullah et al. [66] | Semantic segmentation / DeepLabv3+ | RGB image | Indian Driving Dataset (IDD) Road segmentation mIoU = 98.42% Pothole segmentation mIoU = 73.83% Karlsruhe Institute of Technology and Toyota Technological Institute (KITTI) Dataset Road segmentation F1 score = 95.21% |
|---|---|---|---|
| Fan et al. [67] | Semantic segmentation / DeepLabv3+ | RGB image transformed disparity image | RGB image mIoU = 61.51% mFsc = 76.16% Transformed disparity images mIoU = 72.75 mFsc = 84.22 |
| Fan et al. [68] | Semantic segmentation / DCNNs with GAL | RGB image, disparity image, transformed disparity image | Cityscape GAL-GSCNN mFsc =88.92% IoU=80.186% GAL-DEEPLABV3+ mFsc=62.962% IoU=45.934% ADE20k GAL-GSCNN mFsc = 89.903% IoU = 81.537% GAL-DEEPLABV3+ mFsc = 64.12% IoU = 47.291% |
| Jahanshahi et al. [69] | Hybrid (Classical image processing & 3-D point cloud modelling and segmentation) | 3-D point cloud, Depth image. | Accuracy = 87.9% Precision = 88.1% Recall = 83.9% |
| Fan et al. [70] | Hybrid (Classical image processing & 3-D point cloud | 3-D point cloud, transformed disparity image | Detection accuracy = 98.7% Pixel-level accuracy = 99.6%. |
| | modelling and segmentation) | | |
| Azhar et al. [71] | Hybrid (Classical image processing & machine/deep learning) / Naïve Bayes classifier | RGB Image | Accuracy = 90 % |
| Yousaf et al. [72] | Hybrid (Classical image processing & machine/deep learning) | RGB Image. | Accuracy of pothole detection = 95.7% Accuracy for localization of potholes = 91.4% |
| Anand et al. [73] | Hybrid (Classical image processing & machine/deep learning) | RGB Image. | Precision     0.9237 Recall         0.9376 F1 score 0.9301 |
| Wu et al. [74] | Hybrid (Machine/deep learning & 3-D point cloud modeling and segmentation) | 3-D point cloud, RGB image | The results show that the mean size accuracy is ~1.5–2.7 cm. |

## 3. Proposed Methodology

### 3.1. Data Acquisition

Figure 4 summarizes the overall methodology adopted in this research. The first step was the collection of data. Images were collected from online sources [73] and the pavement in Islamabad, Rawalpindi, and Taxila region.

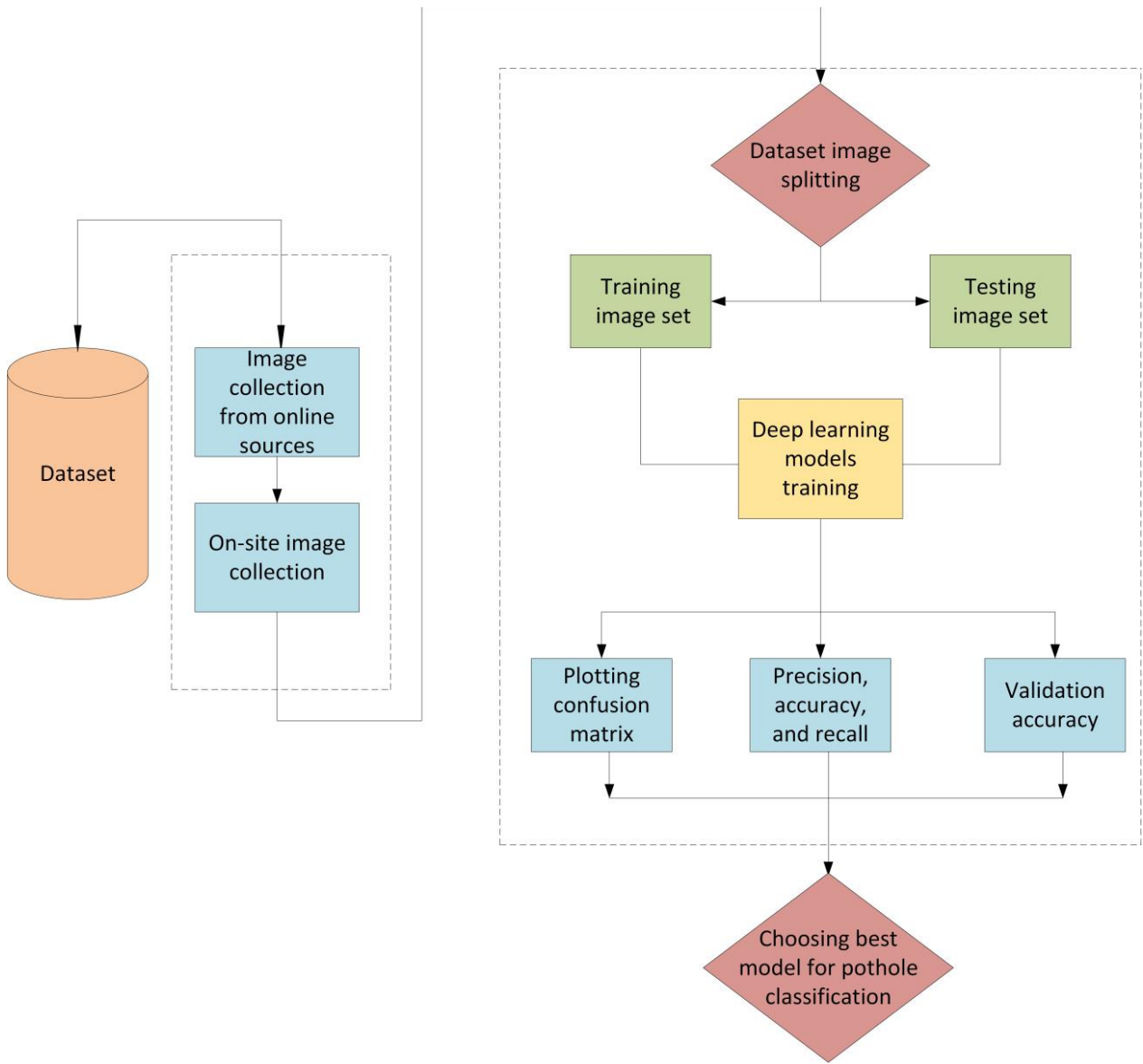

**Figure 4:** Proposed methodology block diagram of potholes detection

For the first classification, data from an online source and images taken from the site were combined, and then the dataset was divided into two classes, Normal and Pothole. As shown in Figure 5, each class has 1000 images for training and 150 for testing.

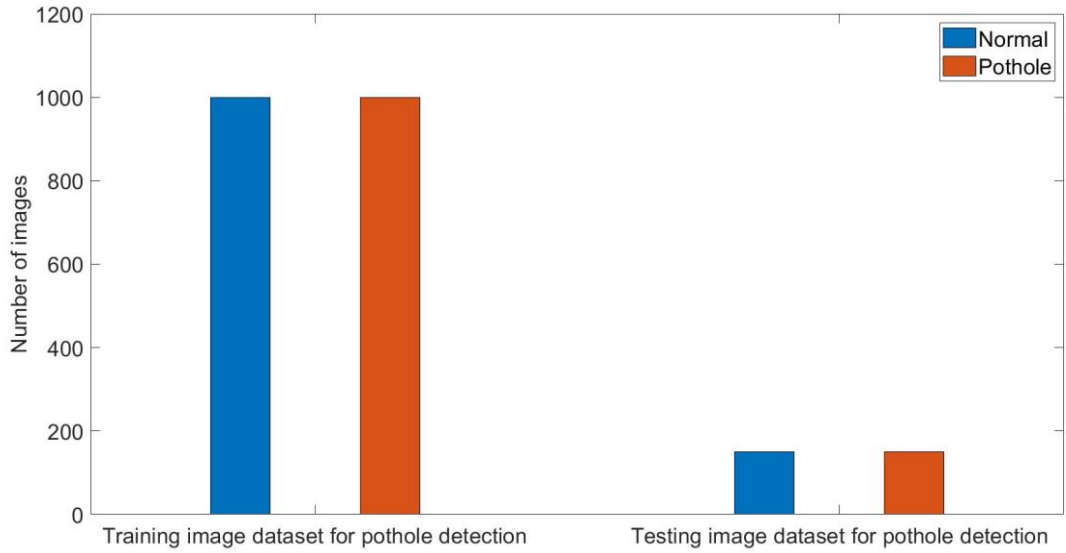

**Figure 5:** Image dataset division for first classification

Images from the site were used exclusively for the second and third classifications. Images were captured at two heights, 2 feet, and 3.5 feet, and subsequently categorized as Normal, Small, and Large. The models were evaluated independently for each height. Figures 6 and 7 represent the height-based image data divisions.

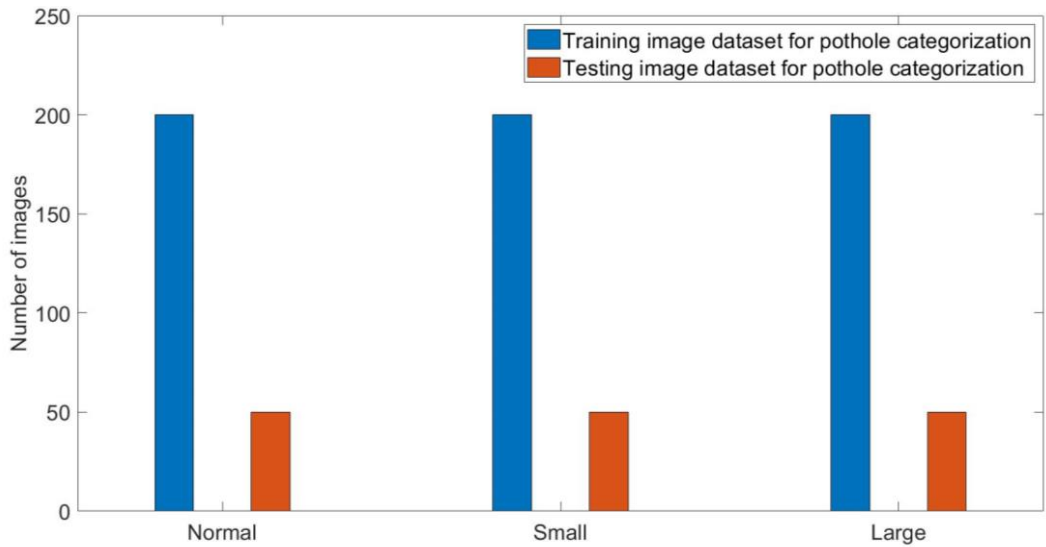

**Figure 6:** Image dataset division for second classification. (2ft)

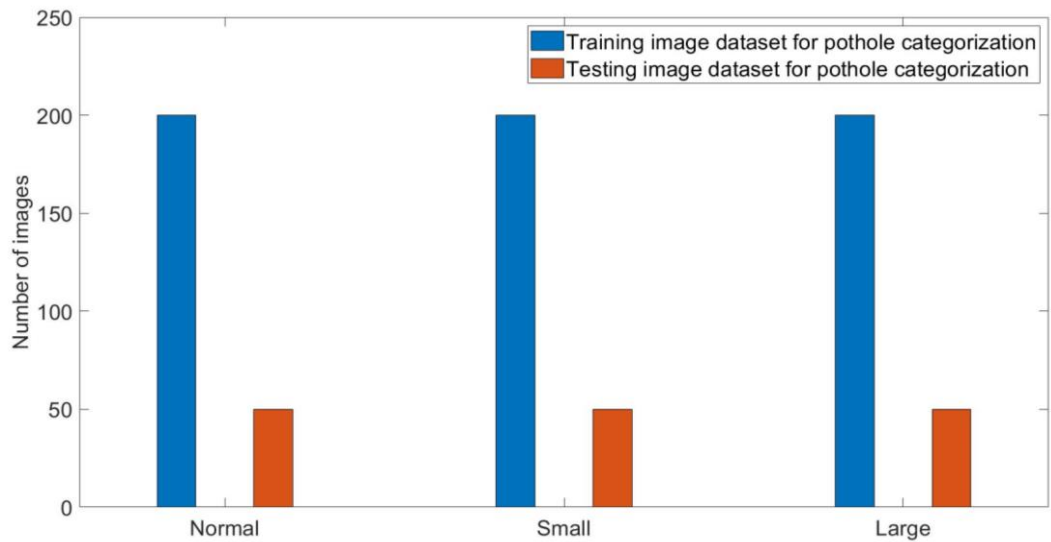

**Figure 7:** Image dataset division for third classification (FFW)

A few of the pothole images taken from online sources are shown in Figure 8. To increase the accuracy of the trained models, image dataset, online images and images taken from the site were used for the training models to detect potholes. These portholes are classified on the basis of size not the failure type.

While Figures 9 and 10 show images taken with mobile phones at 2- and 3.5-feet height, respectively. Images in figure 9 show that the size of the pothole seems to be greater compared to the images in the figure, in fact the size of the pothole was the same. The only change is the height from which the images were taken.

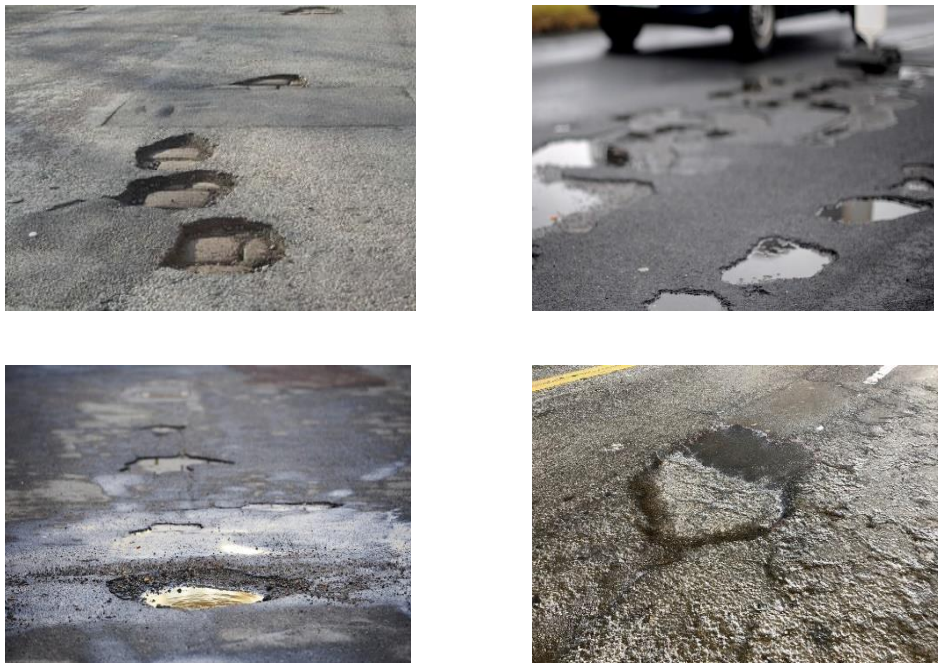

**Figure 8:** Sample pothole images from an online source

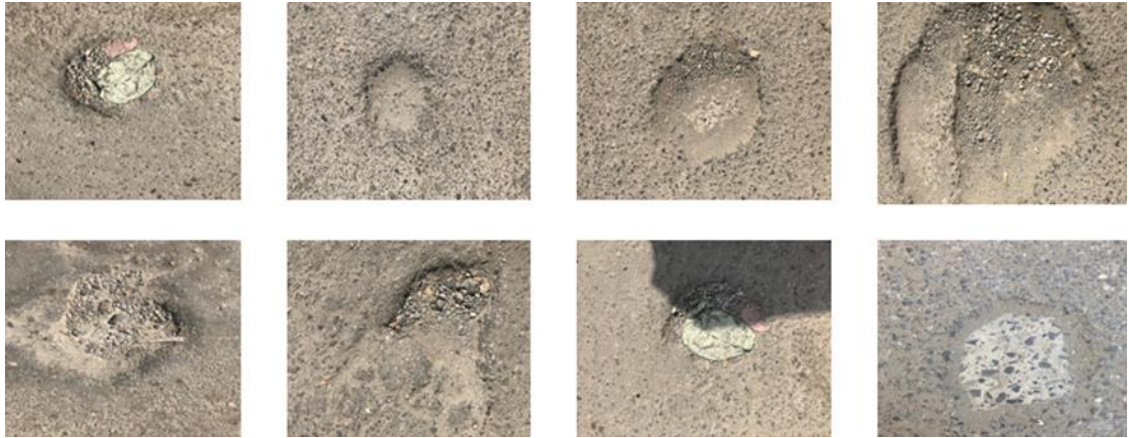

**Figure 9:** Images Captured from 2ft Height

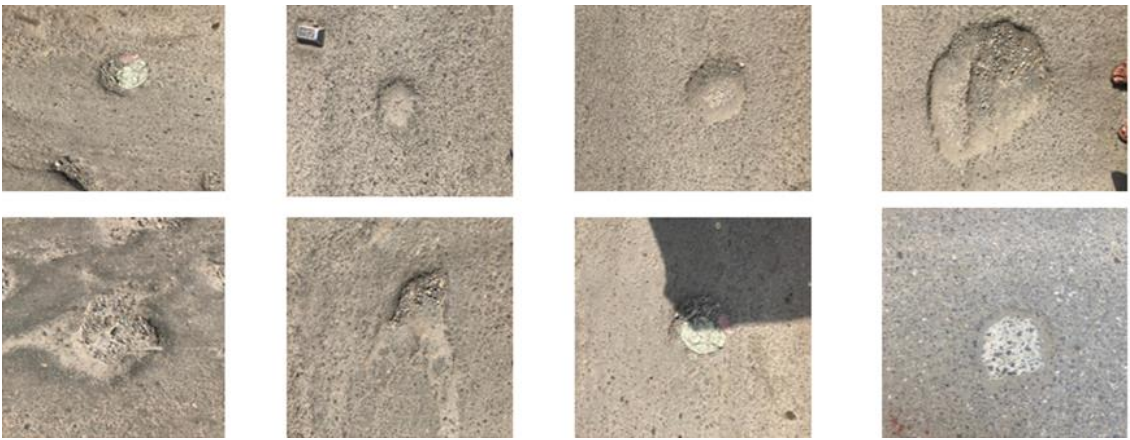

**Figure 10:** Potholes images captured from waist Height- FFW (3.5ft)

### 3.2. Dataset division criteria and Selection of Epochs

In the context of neural networks, an epoch refers to a complete iteration or pass through the entire dataset during the training process. Each epoch involves presenting all the training samples to the network, calculating the loss, and adjusting the model's weights to improve its performance. It essentially captures the point at which the network has had an opportunity to learn from the entire dataset [76].

The selection of the appropriate epoch value is a critical decision in training a neural network. It determines how many times the network will cycle through the dataset, impacting the convergence and generalization of the model. To thoroughly evaluate and determine the optimal epoch value, an empirical study was conducted using the ResNet 50 architecture. This study involved training ResNet 50 with varying epoch values, specifically 1, 2, 3, 4, and 5.

Subsequently, the impact of these different epoch values on the network's performance was assessed. The evaluation was based on the detection of potholes, and a key metric of interest was the validation accuracy. This accuracy reflects the network's ability to generalize and perform well on unseen data. Figure 8 illustrates the relationship between the number of epochs and the corresponding validation accuracy.

Upon analysis of the graph, it becomes evident that as the number of epochs increases, the validation accuracy also improves. However, after a certain point, the validation accuracy plateaus or might even degrade, indicating overfitting. From the graph, it is discernible that an epoch value of 5 yields a validation accuracy consistently surpassing 95%. Consequently, this value was determined to be optimal for achieving the desired performance for the detection of potholes.

Therefore, based on the empirical findings and the observed trend in the graph, an epoch value of 5 was selected as the most suitable choice for training all the chosen convolutional neural network (CNN) architectures.

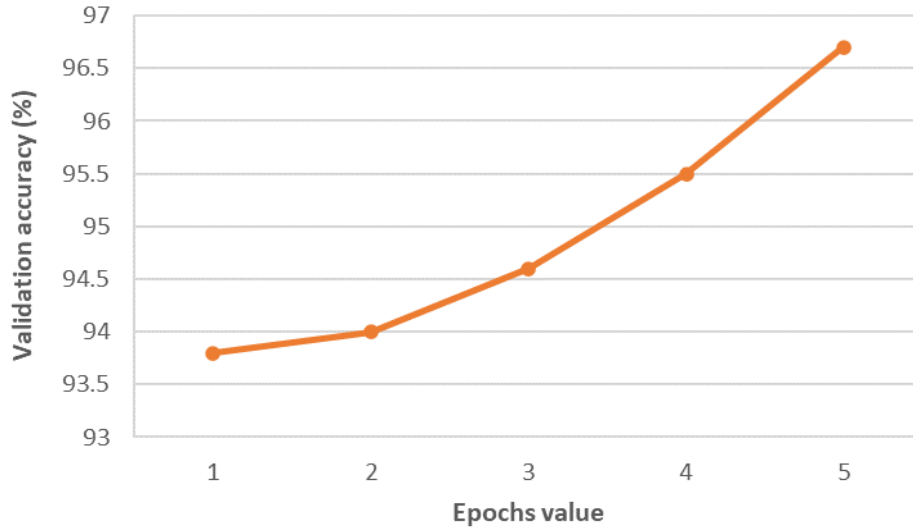

**Figure 11:** Effect of epoch value on validation accuracy for ResNet50

The categorization process involved distinguishing between regular pavement and potholes. In the first classification, no image processing was required to identify the pothole area since both normal and pothole images were easily distinguishable. However, in the second and third classification, the small and large potholes categories respectively, image processing methods were used to identify the contact area. Table 1 specifies that any pothole area larger than 60,000 mm2 falls into the large category.

To determine the area covered by a single pixel, an image of a standard A4 (210 x 270mm) page was taken from a height of 2 ft. and FFW, and the formula (Equation 1) was used.

$$Area\ cover\ by\ single\ pixel\ (pi) = \frac{Area\ of\ A4\ page}{Area\ cover\ by\ A4\ page\ in\ pixels} \qquad (1)$$

The area covered by a single pixel was calculated by dividing the area of the A4 page by the area covered by the A4 page in pixels, which was found using image processing. The area covered by a single pixel was found to be 0.2013 mm2/pixel for the image taken from 2ft and 0.5087 mm2/pixel for the FFW image. Pre- and post-processed versions of the image taken from 2ft with an A4 page are shown in Figures 12(a) and (b).

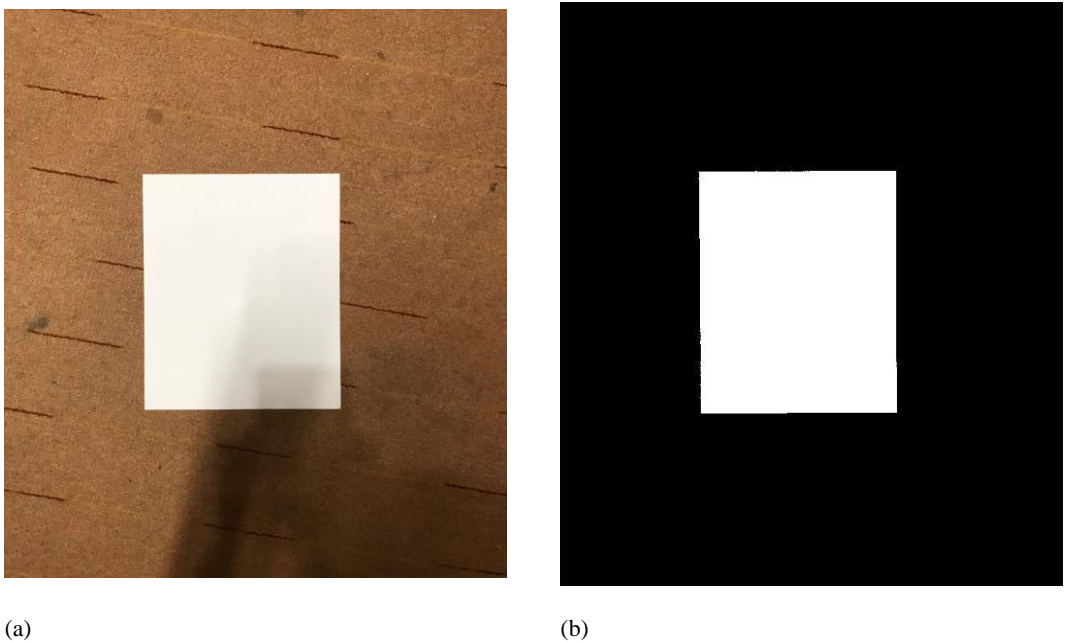

(a)                                                        (b)

**Figure 12:** Image processing for image taken from 2ft.

Figures 13(a) and (b) show the pre- and post-processed versions of the pothole image. The area of the pothole was determined by counting the number of white pixels in the post-processed image and multiplying that number by the area covered by a single pixel. The resulting area was then placed in the appropriate category.

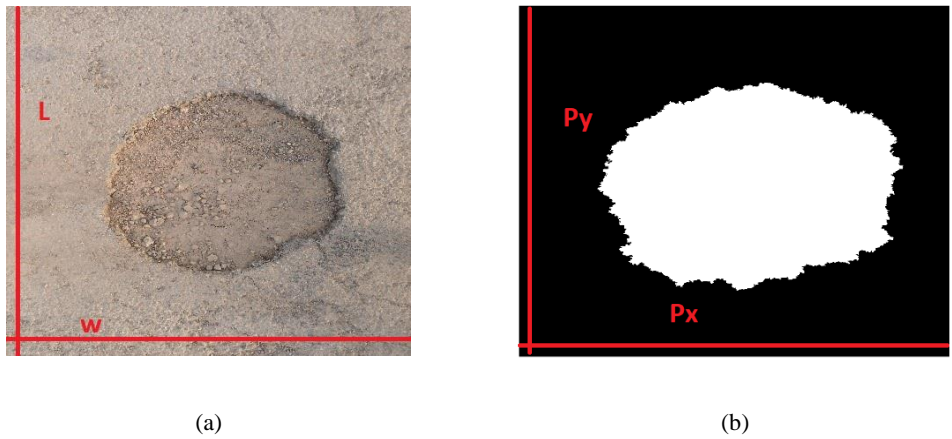

|  (a)  |  (b)  |

**Figure 13:** Samples of image segmentation on potholes

### 3.3. Models Architectures

Three models are selected for the detection and categorization of the pothole. It includes MobileNet v2, ResNet 50, and ResNet 18. MobileNet v2 was selected considering its small size and application in mobile devices. The two variants of ResNet were used because they can be trained easily without increasing the error percentage. ResNet 50 architecture consists of 48 convolutional Layers, 1 MaxPool, and 1 Average pool layer. Figure 14 shows the architecture of ResNet 50.

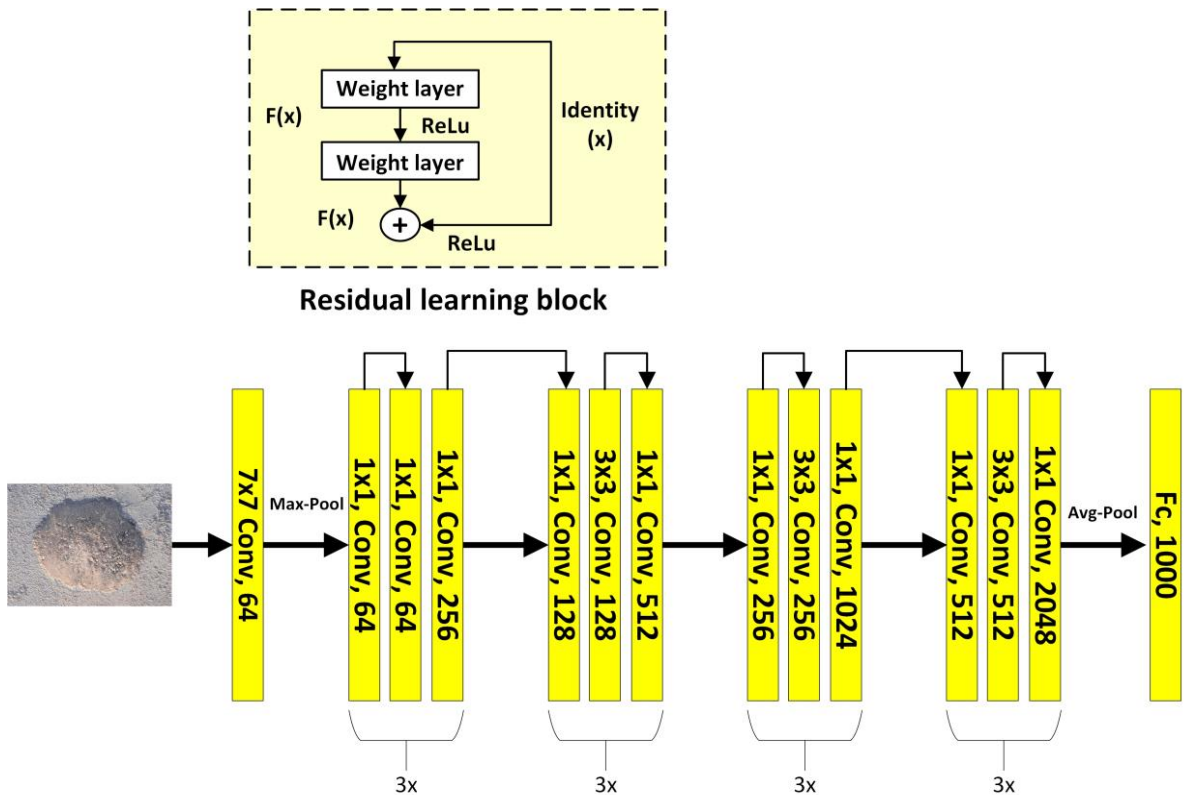

**Figure 14:** Architecture of ResNet 50

Another variant of the ResNet model is ResNet 18. It consists of 72 layers with 18 deep layers. The architecture of ResNet 18 is shown in Figure 15. The figure shows that ResNet 18 skips a few connections, making it faster than ResNet 50 for the model's training.

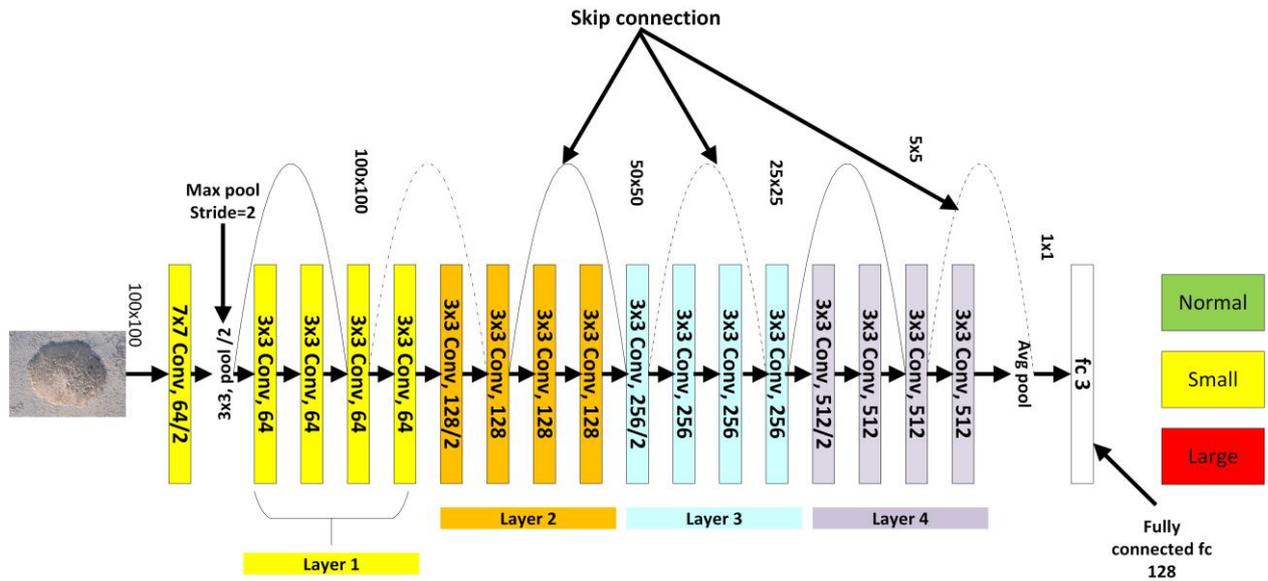

**Figure 15:** Architecture of ResNet 18

MobileNet v2 is designed for mobile and embedded vision applications. Each of its layers consists of depth wise separable convolution. Global average pooling is also used. It consists of 28 deep layers containing both depth wise and pointwise convolutions. The architecture of MobileNet v2 is shown in Figure 16. MobileNet v2, ResNet 18, and ResNet 50 has an image input size of 224 × 224 × 3.

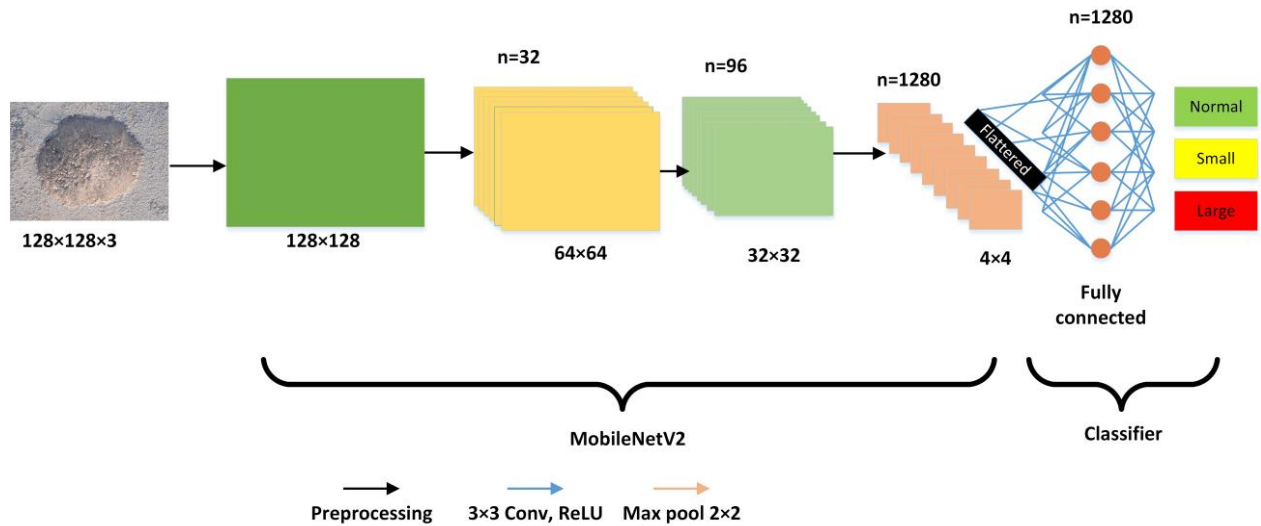

**Figure 16:** Architecture of MobileNet V2

Figures 17, 18, and 19 show losses vs iteration curves for the training of training of three models for the three type of classification. In these figures, we can see that increase in number of epochs and iterations the loss function decreases.

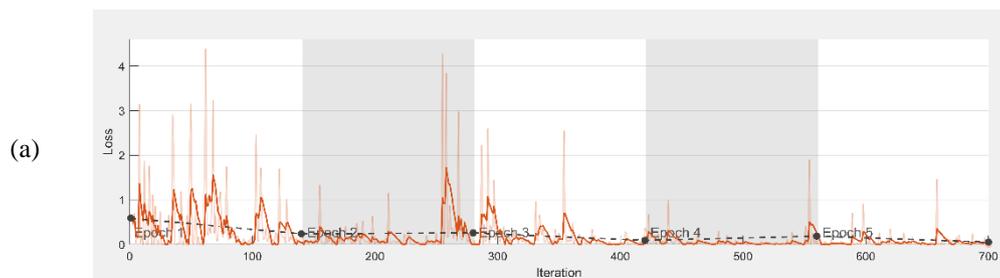

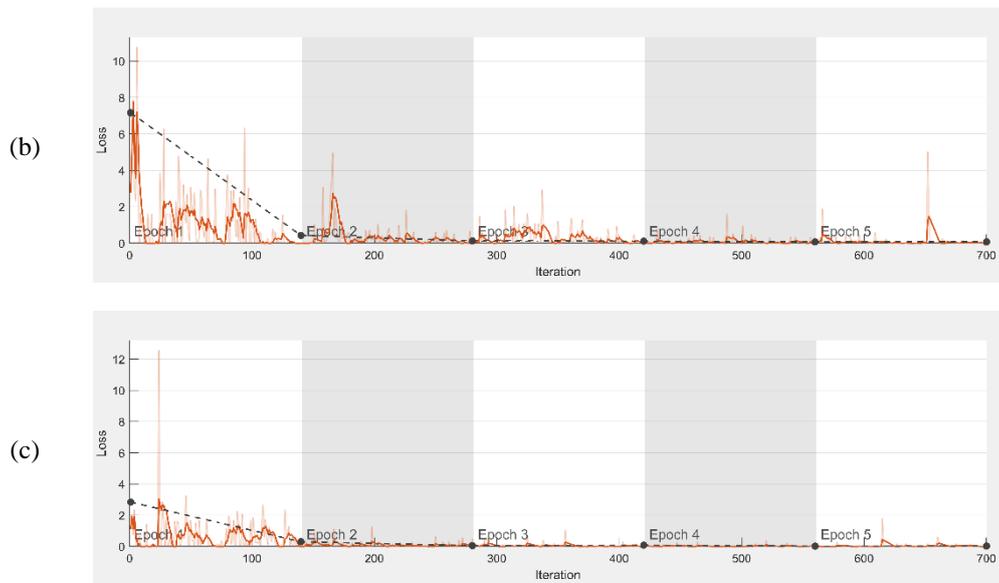

**Figure 17:** Loss vs. iteration curve for first classification (a) MobileNet v2 (b) ResNet 18 (c) ResNet 50

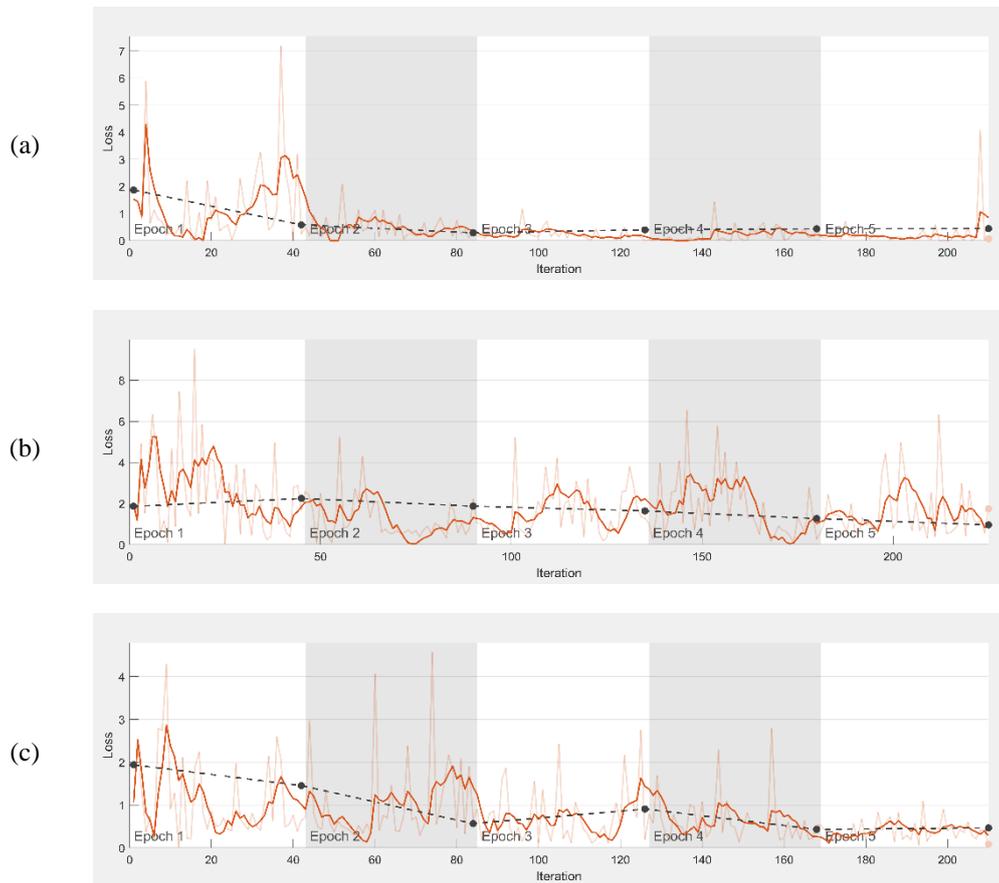

**Figure 18:** Loss vs. iteration curve for second classification (a) MobileNet v2 (b) ResNet 18 (c) ResNet 50

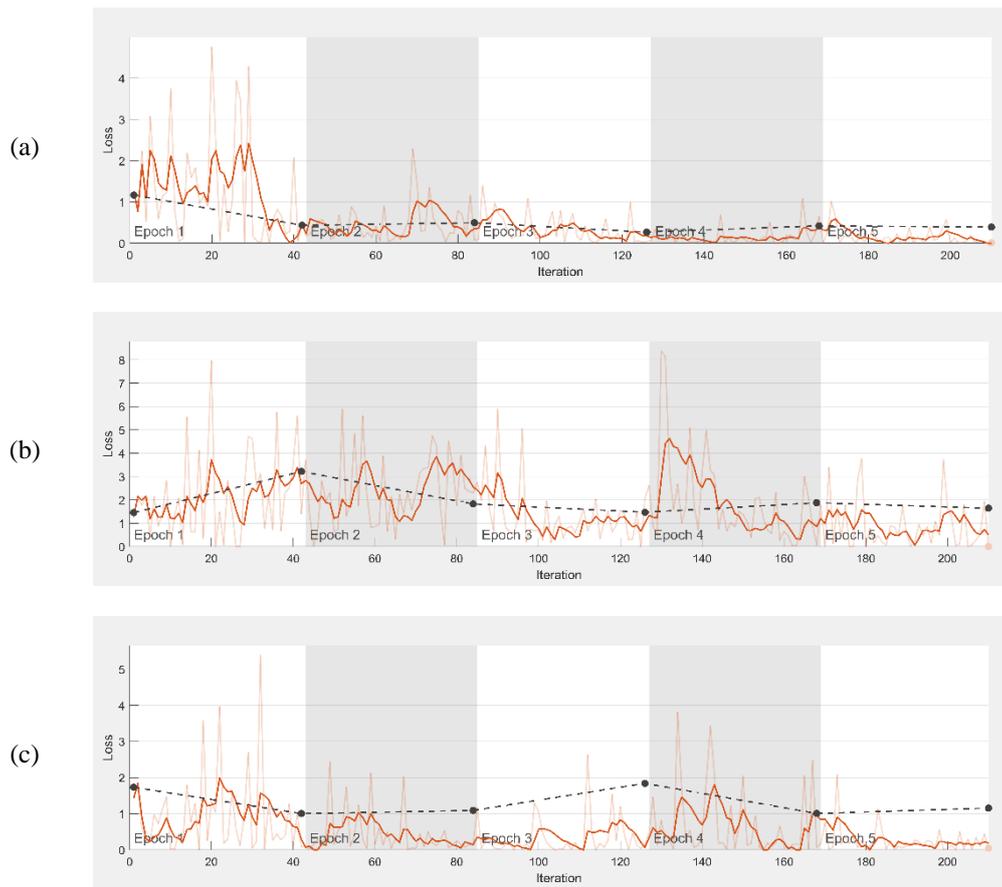

**Figure 19:** Loss vs. iteration curve for third classification (a) MobileNet v2 (b) ResNet 18 (c) ResNet 50

## 4. Experimentation and Results

### 4.1. First classification

300 equally divided images in both classes were tested on each trained model for classifying images with potholes and normal pavement. The Confusion Matrix for each CNN model is shown in Figure 20. It shows that all the models rightly identified images with potholes. While out of 150 images, from the normal pavement category 6, 7 and 20 images were wrongly identified by MobileNet v2, ResNet 50, and ResNet 18, respectively.

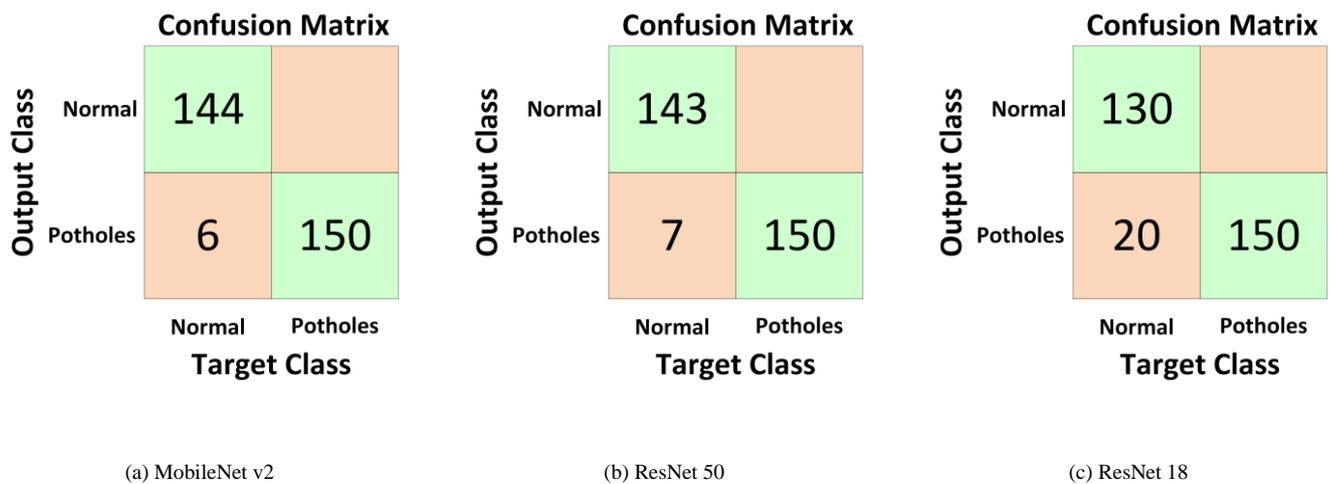

(a) MobileNet v2          (b) ResNet 50          (c) ResNet 18

**Figure 20:** Confusion matrices for first classification

Based on the confusion matrixes shown in Figure 20, the value of accuracy, precision, recall, and f1 score is calculated using equations 2,3,4, and 5, [75] respectively.

$$Accuracy = \frac{TP + TN}{TP + TN + FP + FN} \qquad (2)$$

$$Precision = \frac{TP}{TP + FP} \qquad (3)$$

$$Recall = \frac{TP}{TP + FN} \qquad (4)$$

$$F1\ score = \left( \frac{2 \times Recall \times Precision}{Recall + Precision} \right) \qquad (5)$$

Table 3 summarizes the value of accuracy, precision, recall, f1 score, and training time for each CNN model to detect the pothole.

**Table 3:** Accuracy, precision, F1 score, recall, and training time of CNN models

| Model Used | Accuracy | Precision | F1_Score | Recall | Training Time |
|---|---|---|---|---|---|
| **ResNet50** | 0.9767 | 1.00 | 0.9772 | 0.9554 | 5min 20sec |
| **ResNet18** | 0.9333 | 1.00 | 0.9375 | 0.8824 | 3min 10sec |
| **MobileNet v2** | 0.98 | 1.00 | 0.9804 | 0.9615 | 7min 48sec |

A brief comparison of the accuracy for detecting potholes is shown in Figure 21. Figure 21 indicates that MobileNet v2 performed better in accuracy, with a value equal to 98 %. Precision for classifying potholes is 100% for all the models, indicating that all the images containing potholes are rightly identified. Comparing the training time from Table 3, MobileNet v2 takes more time than the other two, with a value of 7 minutes and 48 sec.

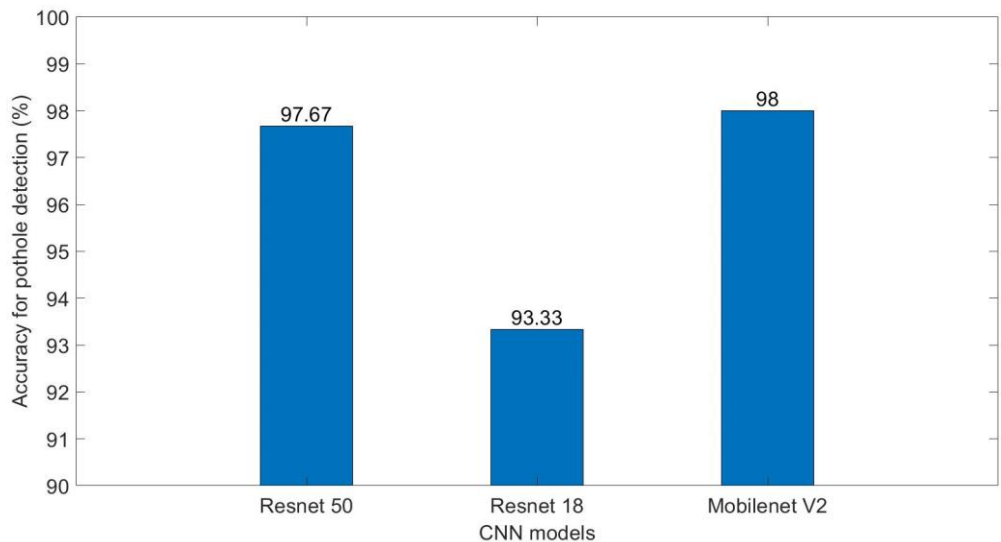

**Figure 21:** Accuracy of models used for first classification

*4.2. Second Classification (Potholes Image taken from 2 feet height)*

150 images subdivided into three classes are tested on trained models for the second classification. The Confusion Matrix for each CNN model trained with images taken from 2 feet height is shown in Figure 22.

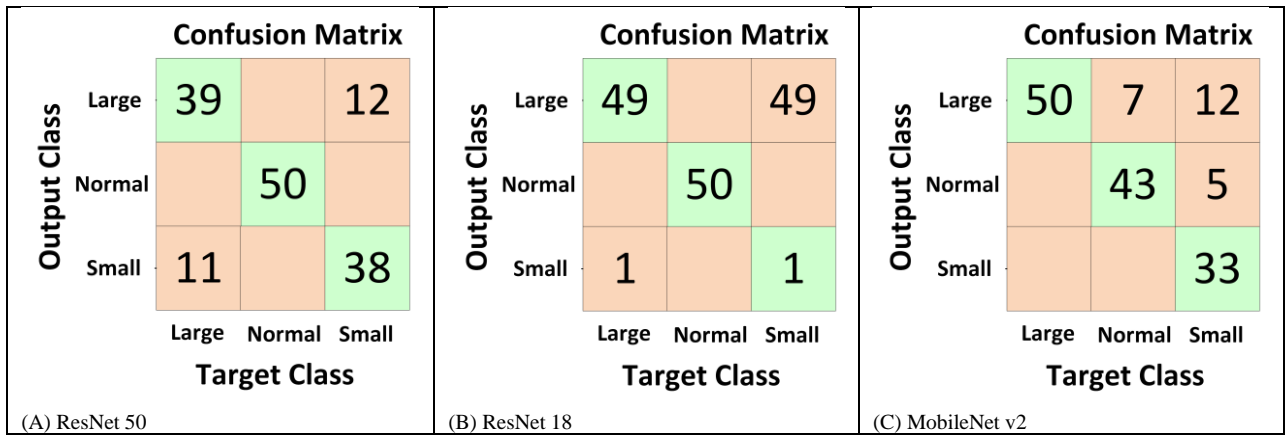

**Figure 22:** Confusion matrix for second classification (A) ResNet 50 (B) ResNet 18 (C) MobileNet v2

Accuracy, precision, recall, f1 score, and training time for each CNN model are in Table 4. Equations 2, 3, 4, and 5 were used to derive the values. The precision of ResNet 50 for classifying large potholes, normal pavement, and small potholes was 76%, 100%, and 78%, respectively. When identifying large potholes, normal pavement, and small potholes, MobileNet v2 was 72%, 90%, and 100% precise, respectively. While for ResNet 18, for the classification of large potholes, normal pavement, and small potholes precision was equal to 50%, 50%, and 100%, respectively.

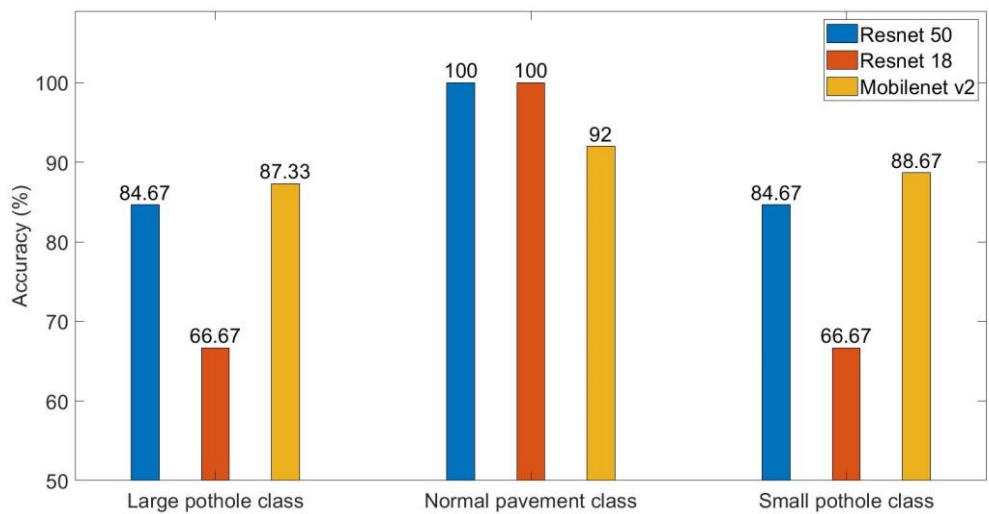

**Figure 23:** Accuracy for different categories (second classification)

Figure 23 illustrates the accuracy of each model in distinguishing between large potholes, small potholes, and normal pavement in 2-foot-high images. MobileNet v2 outperformed the other model with a classification accuracy of 84.67% for large pothole, 92% for normal pavement and 88.67% for small pothole. ResNet 18 and ResNet 50 are equally accurate for classification of the normal pavement with a value of 100%.

*4.3. Third Classification (Potholes Image taken from FFW height):*

150 images subdivided into three classes are tested on trained models for the third classification. The Confusion Matrix for each CNN model trained with the images taken from 3.5 feet is shown in Figure 24.

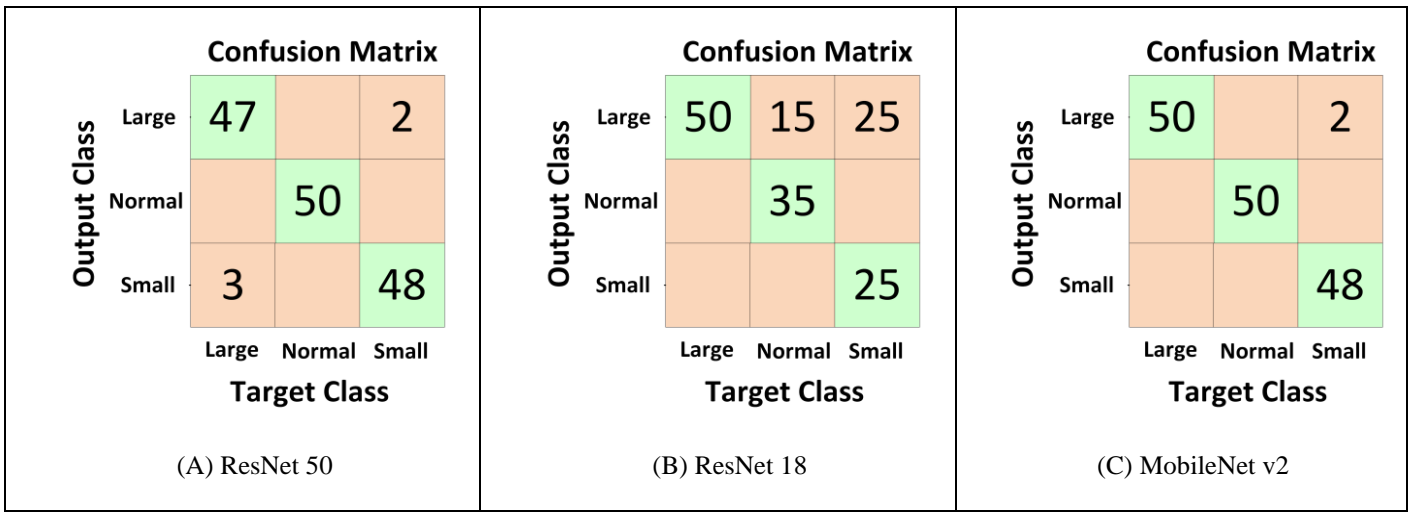

**Figure 24:** Confusion matrix for third classification (FFW)

Table 5 summarizes the value of accuracy, precision, recall, f1 score, and training time for each CNN model. Regarding precision, ResNet 50 has 96%, 96%, and 94% precision for large potholes, normal pavement, and small potholes, respectively. MobileNet v2 has 96%, 100%, and 100% precision for large potholes, normal pavement, and small potholes. ResNet 18 has 56%, 100%, and 100% precision for classifying large potholes, normal pavement, and small potholes. As ResNet18 is a relatively simple model compared to the other two deep neural network architectures. These models have more layers and more complex structures, enabling them to capture more intricate features and patterns in the data. As a result, ResNet18 does not perform as well as these models on more complex tasks. However, the processing time for ResNet 18 is less than the others.

**Table 4:** Accuracy, precision, F1 score, recall, and training time of CNN models for third classification (FFW)

| For Large Category | | | | | |
|---|---|---|---|---|---|
| **Models Used** | **Accuracy** | **Precision** | **Recall** | **F1_Score** | **Training Time** |
| **ResNet50** | 96.67% | 0.96 | 0.94 | 0.95 | 4min 48sec |
| **ResNet18** | 73.33% | 0.56 | 1.0 | 0.71 | 4min 16sec |
| **MobileNet v2** | 98.67% | 0.96 | 1.0 | 0.98 | 5min and 16 sec |
| For Normal Category | | | | | |
| **ResNet50** | 96.67% | 0.96 | 0.94 | 0.95 | 4min 48sec |
| **ResNet18** | 90% | 1.0 | 0.70 | 0.82 | 4min 16sec |
| **MobileNet v2** | 100% | 1.0 | 1 | 1 | 5min and 16 sec |
| For Small Category | | | | | |
| **ResNet50** | 96.67% | 0.94 | 0.96 | 0.95 | 7min 48sec |
| **ResNet18** | 83.33% | 1.0 | 0.50 | 0.67 | 4min 16sec |
| **MobileNet v2** | 98.67% | 1.0 | 0.96 | 0.98 | 5min and 16 sec |

Figure 25 shows each model's accuracy for determining large and small potholes and normal pavement for the images taken at the height of 2 feet. MobileNet v2 performed well compared to the other model, with an accuracy of 98.67 % for classifying large and small potholes. While for normal pavement detection accuracy of MobileNet v2 and ResNet 50 accuracy is equal to 100%.

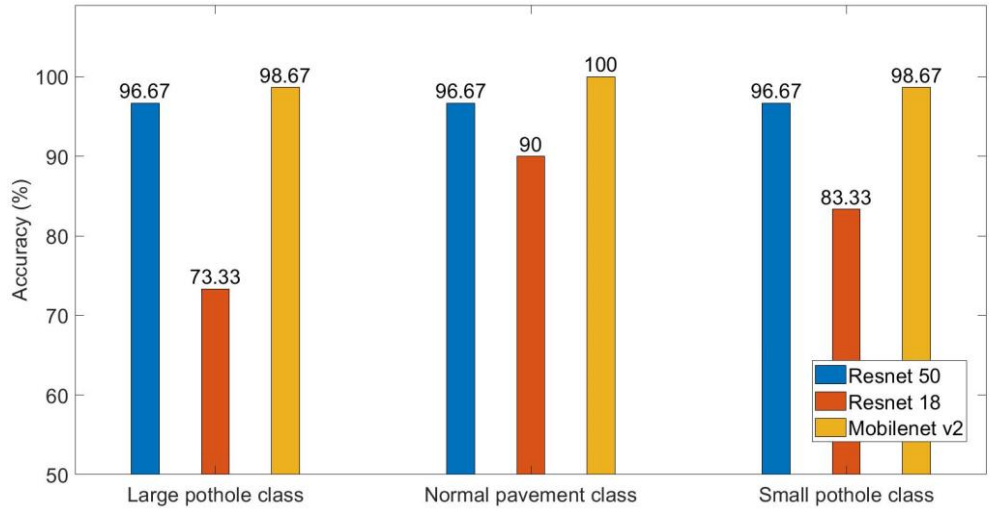

**Figure 25:** Accuracy for different categories (third classification)

Overall, all summaries of the results from sections 4.2 and 4.3 indicate that MobileNet v2 is better for all three types of classification. MobileNet v2 have an accuracy of 98% for sorting pothole in first classification. In the second classification it has an accuracy value of 87.33%, 92% and 88.67% for classifying the large potholes, normal pavement, and small potholes, respectively. Similarly, third classification it has an accuracy value of 98.67%, 100% and 98.67%. The performance of the best suitable model is summarized in Figure 26.

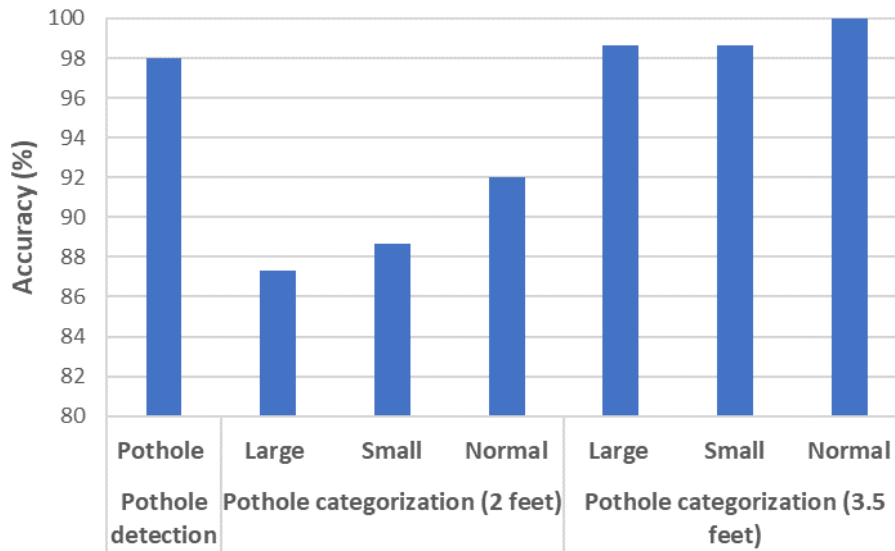

**Figure 26:** MobileNet v2 performance in terms of accuracy

ResNet18 is a relatively simple model compared to the other two deep neural network architectures. These models have more layers and more complex structures, enabling them to capture more intricate features and patterns in the data. As a result, ResNet18 does not perform as well as these models on more complex tasks. However, the processing time for ResNet 18 is less than the others.

## 5. Conclusions

In conclusion, the study aimed to determine the most effective model for automatically classifying pavement images captured from different heights using a mobile phone application or a camera embedded in the car's bumper. Three types of classifications were conducted, which involved normal pavement, small potholes, and large potholes. The MobileNet v2 model was found to be the most efficient model for image classification, despite having a higher training time than ResNet 50 and 18.

The results of the study showed that the MobileNet v2 model achieved a success rate of 98% for identifying potholes from normal pavement. Furthermore, the accuracy values for the classification of images captured at a height of 2 feet were recorded as 87.33 %, 88.67 %, and 92 % for large potholes, small potholes, and normal pavement, respectively. Similarly, the accuracy values for the classification of images captured at a height of 3.5 feet were recorded as 96.7 %, 98.6 %, and 100 % for large potholes, small potholes, and normal pavement, respectively.

Overall, the study demonstrated that the MobileNet v2 model is an effective and reliable method for automatically classifying pavement images. This finding has significant implications for the development of mobile phone applications and car camera systems that can detect and report potholes accurately, thus contributing to the improvement of road safety and maintenance.

**Author Contributions:**

Chauhdary Fazeel Ahmad: Data curation; Visualization; Writing - original draft.

Abdullah Cheema: Visualization; Writing - original draft.

Waqas Qayyum: Data curation; Visualization; Writing - original draft.

Rana Ehtisham: Visualization; Writing - original draft.

Alireza Bahrami: Conceptualization; Formal analysis; Investigation; Methodology; Project administration; Resources; Validation; Writing - original draft; Writing - review & editing.

Muhammad Haroon Yousaf: Formal analysis.

Junaid Mir: Data curation; Formal analysis; Validation.

Afaq Ahmad: Conceptualization; Data curation; Formal analysis; Methodology; Project administration; Resources; Validation.

All authors have read and agreed to the published version of the manuscript.

**Funding:** The authors are thankful to Punjab Higher Education Commission (PHEC) research grant of PHEC/ARA/PIRCA/20529/22 for funding this study.

**Data Availability Statement:** Not applicable.

**Conflicts of Interest:** The authors declare no conflict of interest.